\documentclass[letterpaper, 10 pt, conference]{ieeeconf}  % a4paper
\IEEEoverridecommandlockouts                              % This command is only needed if you want to use the \thanks command
\usepackage{graphicx} %% for loading jpg figures
\usepackage[english]{babel}
\usepackage{amsmath}
\usepackage[mathscr]{euscript}
\usepackage{mathrsfs}
\usepackage{times}
\usepackage[usenames,dvipsnames,svgnames,table, xcdraw]{xcolor} %http://ctan.org/pkg/xcolor
\usepackage{amsfonts}
\usepackage{url}
\usepackage{subcaption}
\usepackage{booktabs}
\usepackage{multirow}
\usepackage[ruled,vlined,linesnumbered]{algorithm2e}
\SetKw{Continue}{continue}
\SetKw{Break}{break}
\let\labelindent\relax
\usepackage{enumitem}

\usepackage{cite} % to group references

\newcommand{\eqcolor}[1]{{{\textcolor{blue}{#1}}}}
\newcommand{\modeldetheuristic}{D-LNS}
\newcommand{\modeldetexact}{D-ES}
\newcommand{\modeluncertain}{S-ES}

\usepackage[breaklinks=true,letterpaper=true,colorlinks,bookmarks=false]{hyperref}

\title{
Human-robot Matching and Routing for Multi-robot Tour Guiding under Time Uncertainty
}

\author{Bo Fu$^{1}$, Tribhi Kathuria$^{1}$, Denise Rizzo$^{2}$, Matthew Castanier$^{2}$, X. Jessie Yang$^{1}$, Maani Ghaffari$^{1}$, and Kira Barton$^{1}$
\thanks{DISTRIBUTION A. Approved for public release; distribution unlimited. (OPSEC 4396)}
\thanks{$^{1}$ Bo Fu, Tribhi Kathuria, X. Jessie Yang, Maani Ghaffari, and Kira Barton are with the University of Michigan, Ann Arbor, MI 48109, USA \{bofu, tribhi, xijyang, maanigj, bartonkl\}@umich.edu}
\thanks{$^{2}$ Denise Rizzo and Matthew Castanier are with the US Army DEVCOM Ground Vehicle Systems Center, Warren, MI 48397, USA \{denise.m.rizzo2.civ, matthew.p.castanier.civ\}@army.mil}
}

\begin{document}
\setlength\abovedisplayskip{5pt}
\setlength\belowdisplayskip{10pt}

\maketitle
\thispagestyle{plain}
\pagestyle{plain}

\begin{abstract}
This work presents a framework for multi-robot tour guidance in a partially known environment with uncertainty, such as a museum.
A simultaneous matching and routing problem (SMRP) is formulated to match the humans with robot guides according to their requested places of interest (POIs) and generate the routes for the robots according to uncertain time estimation.
A large neighborhood search algorithm is developed to efficiently find sub-optimal low-cost solutions for the SMRP.
The scalability and optimality of the multi-robot planner are evaluated computationally. The largest case tested involves 50 robots, 250 humans, and 50 POIs.
A photo-realistic multi-robot simulation was developed to verify the tour guiding performance in an uncertain indoor environment.
% Results demonstrate that the proposed centralized tour planner is scalable, makes a smooth trade-off in the plans under different environmental constraints, and can lead to robust performance with inaccurate uncertainty estimations (within a certain margin).
\newline Supplementary video: {\url{https://youtu.be/jx1RtK0g6fo}}

\end{abstract}

\section{Introduction}\label{sec:introduction}
\begin{figure*}[t!]
	\centering
	\includegraphics[width=\linewidth]{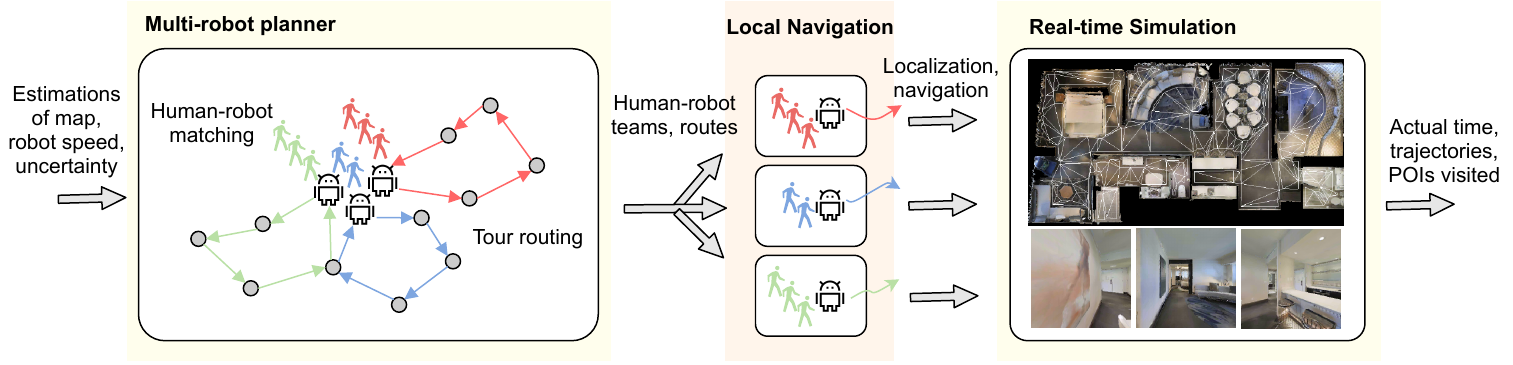}
	\caption{A framework for applying our multi-robot planner within a simulation. A centralized multi-robot tour planner (Sec. \ref{sec:global_planner}) first generates the teams and tour plans for all robots. Then, individual robots use a local navigation planner (Sec. \ref{sec:experiment}) to follow the planned routes and lead the humans within the simulation environment. }
	\label{fig:framework_diagram}
\end{figure*}

Robot tour guides have been applied in different environments \cite{satake2009approach, shiomi2009field, doering2019neural, kirby2010affective, burgard1998interactive, burgard1999museum} because they save human labor, do not require time-consuming training, and have the possibility to ease discomfort during social interactions \cite{de2019social}. Previous works have been focusing on the routing of places of interest \cite{preferences, fu2022learning},
localization and navigation \cite{burgard1998interactive, burgard1999museum},
and social interactions with humans \cite{bennewitz2005towards}.
While they cover diverse topics, these works mainly focus on a single robot guide for a small predetermined human team.
In practice, it is unlikely that a single robot is enough for guiding all human visitors. Therefore, a multi-robot team is a desirable solution. In such a situation, instead of considering each robot separately (as in previous works), a crucial question is how to coordinate the robots as an integrated multi-robot system such that the overall efficiency/performance of the system is maximized.

A natural first step is to consider loose coordination:
once the humans are split into teams and assigned to each robot,
the single-robot tour guide systems in previous works can be applied to the low-level navigation and interaction tasks.
However, the problems of choosing the best robot guides for a human (the matching problem) and finding the optimal tour plan for a robot (the routing problem) are tightly coupled, leading to challenges during the optimization.
This work focuses on a multi-robot planning system that simultaneously optimizes the matching and routing problem.
The optimization planner proposed can be paired with a local navigation/interaction planner to handle practical multi-robot guidance applications.

The actual touring times may vary due to dynamic movements of the teams, changes in the environment, and varying visiting times at a POI. Such variations will be considered as uncertainties in the traveling and visiting time in the optimization to generate time-robust touring plans.

A computational investigation is conducted with the proposed multi-robot planning system. Then, the system performance in an uncertain environment is evaluated through a photo-realistic simulation.
A system diagram of the simulation framework is shown in Fig. \ref{fig:framework_diagram}. 

This work has the following contributions.
\begin{enumerate}[label={\arabic*)}]
    \item The algorithmic modeling of a \textbf{simultaneous} human-robot \textbf{matching} and tour \textbf{routing problem (SMRP)} for multi-robot tour guidance under execution time uncertainty.
    \item A comprehensive computational evaluation of the scalability and solution quality of the proposed algorithms.
    \item The simulation verification of the proposed multi-robot system through a concrete tour guiding case study in a photo-realistic indoor environment.
\end{enumerate}

\section{Related Work}\label{sec:related_work}

The SMRP combines two classic problem types from the operations research: bipartite matching and vehicle routing.

The matching problem refers to splitting the humans into teams such that the matching between humans and robots minimizes the number of dropped human requests. It is a bipartite matching problem as there are two separate sets of elements (robots and humans) in the problem.
Optimal solutions can be found through linear programming or maximum flow optimization \cite{ford2015flows} in polynomial times. Bipartite matching has been applied to multiple real-world problems including image feature matching \cite{cheng1996maximum}, object detection \cite{carion2020end}, budget allocation \cite{aggarwal2011online}, and task assignment \cite{dutta2019one}.

The vehicle routing problem (VRP) considers the minimum traversing distance of all the places of interest using multiple vehicles (robots), which is NP-hard to solve optimally.
The stochastic vehicle routing problem (SVRP) is a variation of VRP where some of the parameters in the optimization are random distributions.
Previous works in the field of SVRP have investigated optimization under uncertain requests number at each POI \cite{mendoza2013multi,secomandi2009reoptimization,marinakis2013particle, fu2022robust}, 
uncertain time costs for traveling edges or service at a POI \cite{sundar2017path,chen2014optimizing,li2010vehicle,gomez2016modeling},
and uncertain energy costs \cite{venkatachalam2019two,venkatachalam2018two, fu2020heterogeneous}. This work considers the uncertainty in time costs.

The SMRP problem in this paper contains multiple vehicle routing problems (formally, a VRP can be reduced to an SMRP); and since a VRP is NP-Hard \cite{toth2002vehicle}, an SMRP is NP-hard.
A ride-sharing system can be regarded as an SMRP where the riders should be matched with drivers while the routes are determined simultaneously. However, many systems in previous work decouple the problems by generating the vehicle routes first and then matching humans with the closest routes \cite{schreieck2016matching, aydin2020matching}.
Unlike previous work, we directly address this joint optimization problem, model it as a mixed-integer optimization, and provide methods that efficiently find sub-optimal low-cost solutions in Sec. \ref{sec:global_planner}.

\section{Human-robot Matching and Tour Routing}\label{sec:global_planner}

This section defines the simultaneous matching and routing problem for multi-robot tour guidance, encodes it as a mixed-integer program, and provides a heuristic algorithm.

\subsection{Problem Description and Graphical Model}

Suppose there is a set of robots \(V = \{1, \cdots, n_V\}\), humans \(L = \{1, \cdots, n_L\}\), and places of interest (POIs) \(M = \{1, \cdots, n_M\}\). Each human can request a subset of POIs to visit.
In this problem, a planner needs to determine which robot a human should follow and what routes (a sequence of POIs) a robot should take, such that the number of satisfied human requests (POIs) is maximized within certain touring time limits.

\begin{figure}[t!]
	\centering
	\includegraphics[width=0.53\linewidth]{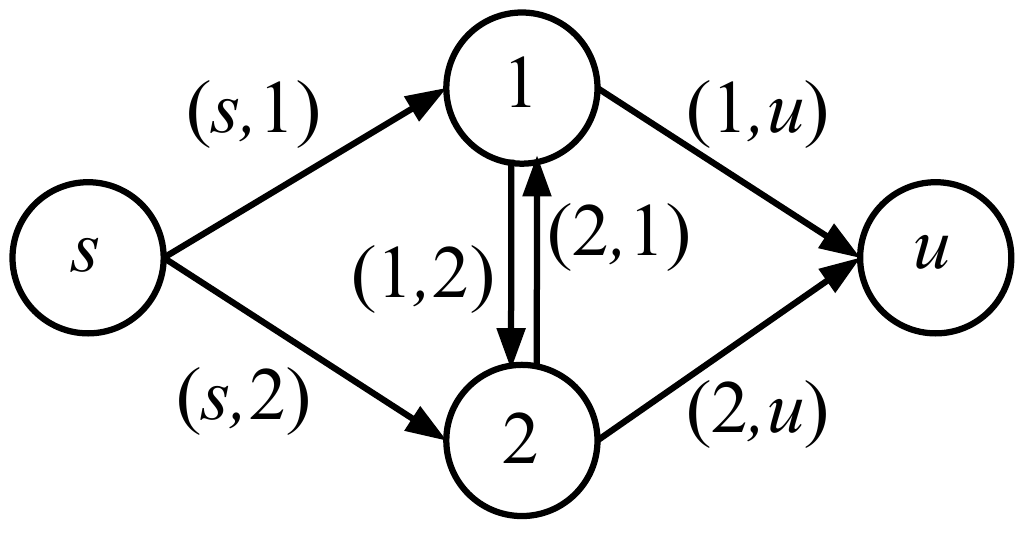}
	\caption{A graphical model example with two POIs. \(s\) and \(u\) are the start and terminal, respectively.}
	\label{fig:graphical_model}
\end{figure}

Suppose the start and terminal locations of the robots are nodes (places) \(s = n_M + 1\) and \(u = n_M + 2\) in the graph. Let \(N = \{1, \cdots, n_M, s, u\}\) be the whole set of places.
We first define a directed graph \(G = (N, E)\), with \(N\) and \(E\) the sets of nodes and edges, respectively (see Figure \ref{fig:graphical_model}).
Note that edge set \(E = \{(i,j)\}, \ \forall i,j \in N\) s.t. \(i \neq j\).

\subsection{A Mixed-integer Bilinear Program}\label{sec:deterministic_math}

\begin{figure}[t!]
	\centering
	\includegraphics[width=0.92\linewidth]{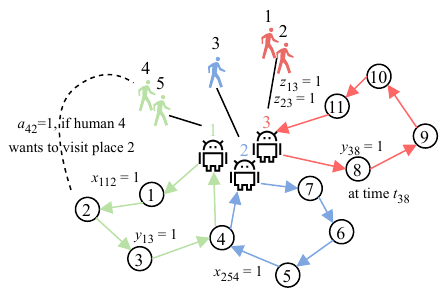}
	\caption{An illustration of the key notations through an example matching and routing.}
	\label{fig:framework_model}
\end{figure}

Here we discuss a mixed-integer bilinear program that mathematically models the SMRP we discussed above. We first provide the notations that will be used in Table \ref{tab:variable_definition}.
The decision variables are \(x_{kij}\), \(y_{ki}\), \(t_{k}\), and \(z_{lk}\).
A graphical illustration of the key notations is shown in Figure \ref{fig:framework_model}.

\begin{table}[ht]
  \caption{Definition of the notations. \(T_{k i}\) and \(T_{k i j}\) are modeled as random variables to capture the uncertainty.}
  \label{tab:variable_definition}
    \small
    \begin{tabular}{p{0.06\linewidth}|p{0.8\linewidth}} 
    \toprule
     & Meaning
    \\
    \midrule
    \(x_{kij}\) & = 1, if robot \(k \in V\) travels \((i,j) \in E\), otherwise 0. \\
    \(y_{ki}\) & = 1, if robot \(k \in V\) visits node \(i \in M\), otherwise 0. \\
    \(t_{k}\) & The time robot \(k \in V\) complete the tour. \\
    \(z_{lk}\) & = 1, if human \(l \in L\) is assigned to follow robot \(k \in V\), otherwise 0. \\
    \(a_{li}\) & = 1, if human \(l \in L\) wants to visit node \(i \in M\), otherwise 0. \\
    \(T_{k i}\) & The time for robot \(k\) and its team to visit POI \(i\). \\
    \(T_{k i j}\) & The time for robot \(k\) and its team to travel edge \((i,j)\). \\
    \(T_k^\mathrm{lim}\) & A preset time limit for the tour guided by robot \(k\). \\
    \(Z_k^{\max}\) & The maximum number of humans guided by robot \(k\). \\
    \bottomrule
    \end{tabular}
\end{table}

\noindent\textbf{Objective Function:}
The objective function \eqref{eqn:bilinear_objective} contains the weighted combination of dropped human requests and time usage, with the weights \(C_a\) and \(C_t\) (\(C_a >> C_t\)), respectively.
The left part is the total number of dropped POIs for all humans.
The right part places a penalty on the scenarios when the tour time is close to the time limit \(T_k^\mathrm{lim}\), with \(\Delta T_k^\mathrm{lim}\) as the margin.
\([x]^+ = x\), if \(x \geq 0\); otherwise, \([x]^+ = 0\).
Such a penalty model has shown effectiveness in previous work in VRP with stochastic time cost \cite{sundar2017path} to ensure time limits under uncertainty. \(\mathbb{E}(\cdot)\) takes the expectation of a variable.
\begin{align}
    \min \
    C_a \cdot \sum_{l \in L} \sum_{k \in V} z_{l k}\left(\sum_{i \in M} a_{l i} \cdot\left(1-y_{k i}\right)\right) \label{eqn:bilinear_objective} \\
    + C_t \sum_{k \in V} \mathbb{E}\left( [t_{k} - (T_k^\mathrm{lim} - \Delta T_k^\mathrm{lim}) ]^+\right). \nonumber
\end{align}

\noindent\textbf{Variable Bounds:}
The feasible regions and bounds of four sets of decision variables are defined as in \eqref{eqn:var_bounds}.
\begin{align}
    x_{k i j}, \ y_{k i}, \ z_{l k} \in \{0,1\}, \quad t_{k} \geq 0, \label{eqn:var_bounds} \\
    \forall k \in V, \ \forall i, j \in N, \ \forall l \in L. \nonumber
\end{align}

\noindent\textbf{Network Flow Constraints:}
Equation \eqref{eqn:flow_constraints1} is a network flow constraint that ensures the incoming robot number equals the outgoing robot number. Constraint \eqref{eqn:flow_constraints2} ensures that there is only one path in the network flow of robot \(k\). Constraint \eqref{eqn:var_relation_constraint2} shows the relationship between variables \(x\) and \(y\).
\begin{align}
    \sum_{i \in N} x_{k i m}=\sum_{j \in N} x_{k m j}, \quad & \forall m \in M, \quad \forall k \in V. \label{eqn:flow_constraints1} \\
    \sum_{i \in N} x_{k s i} \leq 1, \quad \quad \quad & \forall k \in V.  \label{eqn:flow_constraints2} \\
    y_{k j} = \sum_{i \in N} x_{k i j}, \quad \quad & \forall j \in M, \quad \forall k \in V. \label{eqn:var_relation_constraint2}
\end{align}

\noindent\textbf{Time Constraints:}
The total tour time \(t_{k}\) equals the sum of all traveling time and visiting time as in \eqref{eqn:sum_tour_time}. \(t_{k}\) is a random variable as the traveling time \(T_{kij}\) and visiting time \(T_{ki}\) are modeled as random variables (uncertain).
Constraint \eqref{eqn:time_limit_constraint} sets a time limit \(T_k^\mathrm{lim}\) for the whole tour.
\begin{align}
    & t_{k} = \sum_{i \in N} \sum_{j \in N} T_{kij} \cdot x_{kij} + \sum_{i \in N} T_{ki} \cdot x_{ki}, \quad \forall k \in V. \label{eqn:sum_tour_time} \\
    & \mathbb{E}( t_{k} ) \leq T_k^\mathrm{lim}, \quad \quad \quad \quad \quad \quad \quad \quad \quad \quad \quad \ \ \forall k \in V. \label{eqn:time_limit_constraint}
\end{align}

\noindent\textbf{Team Size Limit Constraints:}
All humans must be matched to a robot guide, introducing the constraint \eqref{eqn:visitor_assignment_constraint}.
We also impose a limit to the maximum number of humans that follow a robot to avoid imbalanced teams as in \eqref{eqn:team_size_constraint}. This limit is denoted as \(Z_k^{\max}\).
\begin{align}
    \sum_{k \in V} z_{l k} = 1, \quad \quad \quad \quad & \forall l \in L. \label{eqn:visitor_assignment_constraint} \\
    \sum_{l \in L} z_{l k} \leq Z_k^{\max}, \quad \quad \quad \quad & \forall k \in V. \label{eqn:team_size_constraint}
\end{align}

\subsection{A Large Neighborhood Search Approach}\label{sec:large_neiborhood_search}

Global optimal solutions can be obtained through a branch-and-cut algorithm using existing exact solvers (e.g., the GUROBI solver). However, the computational cost grows exponentially with the problem size.
Here, we propose a large neighborhood search (LNS) approach to provide suboptimal solutions for large problem cases \cite{chen2014optimizing, fu2021simultaneous, he2018improved, alinaghian2018multi, deng2021room, deng2023learning}.
The LNS iteratively solves the SMRP within the large neighborhoods of the two sub-problems as in Algorithm \ref{alg:large_neighborhood_search}.

\begin{algorithm}[t]
% \SetAlgoLined
\small

\textbf{Input:} the unsolved optimization problem in \eqref{eqn:bilinear_objective}-\eqref{eqn:team_size_constraint}

Randomly initialize \(x_{kij}\), \(y_{ki}\), \(z_{lk}\), \(t_{k}\)

\For{\textnormal{iteration} \(= 1, \cdots, n_{\max}\)}{
    Fix  \(x_{kij}\), \(y_{ki}\), \(t_{k}\) \(\quad (\forall k \in V, \ \ \forall i,j \in N)\)
    
    Solve the matching sub-problems in \eqref{eqn:matching_subproblem}
    
    Update \(z_{lk}\)  \(\quad (\forall k \in V, \ \ \forall l \in L)\)
    
    \For{\(k \in V\)}{
        Fix \(z_{lk}\) \(\quad (\forall l \in L)\)
        
        Solve the routing sub-problems in \eqref{eqn:routing_subproblem2}
        
        Update \(x_{kij}\), \(y_{ki}\), \(t_{k}\) \(\quad (\forall i,j \in N)\)
    }
    
    \If{\textnormal{the objective value does not change}}
    {
        \Break
    }
}

\Return the solution \(x_{kij}\), \(y_{ki}\), \(z_{lk}\), \(t_{k}\)

\caption{Large Neighborhood Search}
\label{alg:large_neighborhood_search}
\end{algorithm}

\textbf{Matching Sub-problem:}
Fix the variables \(x_{kij}, y_{ki}, t_{k}\) and solve the SMRP within the neighborhood of \(z_{lk}\).
The problem reduces to the following (variables in blue).
\begin{align}
    \min & \ \
    \sum_{l \in L} \sum_{k \in V} \eqcolor{z_{l k}} (\sum_{i \in M} a_{l i} \cdot\left(1-y_{k i}\right) ) \label{eqn:matching_subproblem} \\
    \text{sub to} & \ \ \eqcolor{z_{lk}} \in \{0, 1\}
    \text{ and} \ \ \eqref{eqn:visitor_assignment_constraint}-\eqref{eqn:team_size_constraint}. \nonumber
\end{align}

This is a standard bipartite matching problem whose integer solutions can be obtained by solving the reduced linear program (replacing \(z_{lk} \in \{0, 1\}\) with \(0 \leq z_{lk} \leq 1\)) in a polynomial time \cite{heller1956extension}.

\textbf{Routing Sub-problem:}
Fix variable \(z_{lk}\) and solve the SMRP optimization within the neighborhood of \(x_{kij}, y_{ki}, t_{k}\).
Since the human-robot matching is fixed (fixed \(z_{lk}\)), a robot only considers the requests of the people in its own team. The optimization is decoupled into \(n_V\) single-vehicle routing problems, with efficient heuristic solvers.
For the version without uncertainty in the time \(T_{k i j}\) and \(T_{k i}\), we use the Google Or-Tools to encode this model. For the version with uncertainty, we use GUROBI.
\begin{align}
    \min \ & 
    C_a \sum_{l \in L} z_{l k} (\sum_{i \in M} a_{l i} \cdot\left(1- \eqcolor{y_{k i}}\right) ) \nonumber \\
    + & C_t \mathbb{E}\left( [\eqcolor{t_{k}} - (T_k^\mathrm{lim} - \Delta T_k^\mathrm{lim}) ]^+\right).
 \label{eqn:routing_subproblem2} \\
    \text{sub} \   \text{to} \ & \ \eqref{eqn:var_bounds}-\eqref{eqn:time_limit_constraint}. \nonumber
\end{align}

\section{Experiments and Results}\label{sec:experiment}

\subsection{Computational Evaluation of the Tour Planning} \label{sec:computational_experiments}

Randomized test cases are generated to evaluate the scalability and optimality of the proposed problem and algorithm (on a laptop with an Apple M1 chip).
The robot number \(n_V\), human number \(n_L\), and the number of POIs \(n_M\) are chosen from the sets \(n_V \in \{4, 10, 20, 50\}\), \(n_L \in \{10, 50, 100, 250\}\), and \(n_M \in \{10, 20, 30, 50\}\).
Three problem types in Table \ref{tab:algorithm_tested} are evaluated on the cases.

\begin{table}[tbp]
  \centering
  \small
  \caption{Problem types. Stochastic/deterministic: whether there is uncertainty in the time costs \(T_{k i}\) and \(T_{k i j}\).}
    \begin{tabular}{c|l}
    \toprule
    \multicolumn{1}{c|}{Abbr} & \multicolumn{1}{c}{Meaning} \\
    \midrule
    \modeldetheuristic{} & Deterministic formulation with LNS algorithm \\
    \modeldetexact{}     & Deterministic formulation with exact solution method \\
    \modeluncertain{}    & Stochastic formulation with exact solution method \\
    \bottomrule
    \end{tabular}
  \label{tab:algorithm_tested}
\end{table}

Results show that the largest problem where the exact solutions (\modeldetexact{} and \modeluncertain{}) can find a non-trivial solution within 120 seconds, consists of 10 robots, 50 humans, and 50 POIs. A trivial solution is defined as dropping all requests and conducting no tour.
In contrast, \modeldetheuristic{} can return a non-trivial solution for all the cases. For the case with 50 robots, 250 humans, and 50 POIs, the \modeldetheuristic{} completes the optimization within 26.5 seconds.
By comparing the results of  \modeldetheuristic{} and \modeldetexact{}, the LNS algorithm finds lower-cost solutions for 13/16 of the cases.

\begin{figure}
\centering
\includegraphics[width=0.28\linewidth, trim=0 0 0 0, clip]{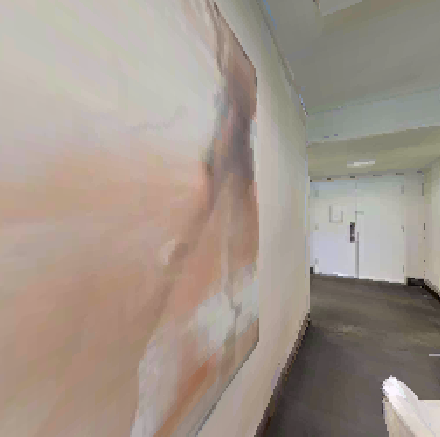}
\includegraphics[width=0.28\linewidth, trim=0 0 0 0, clip]{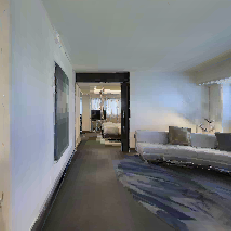}
\includegraphics[width=0.28\linewidth, trim=0 0 0 0, clip]{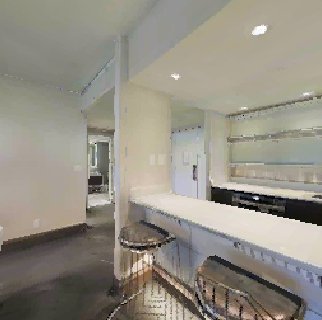}
	\caption{Example views of the robots conducting tasks.}
	\label{fig:camera_view}
\end{figure}

\begin{figure}
    \centering
    \includegraphics[width=0.8\linewidth]{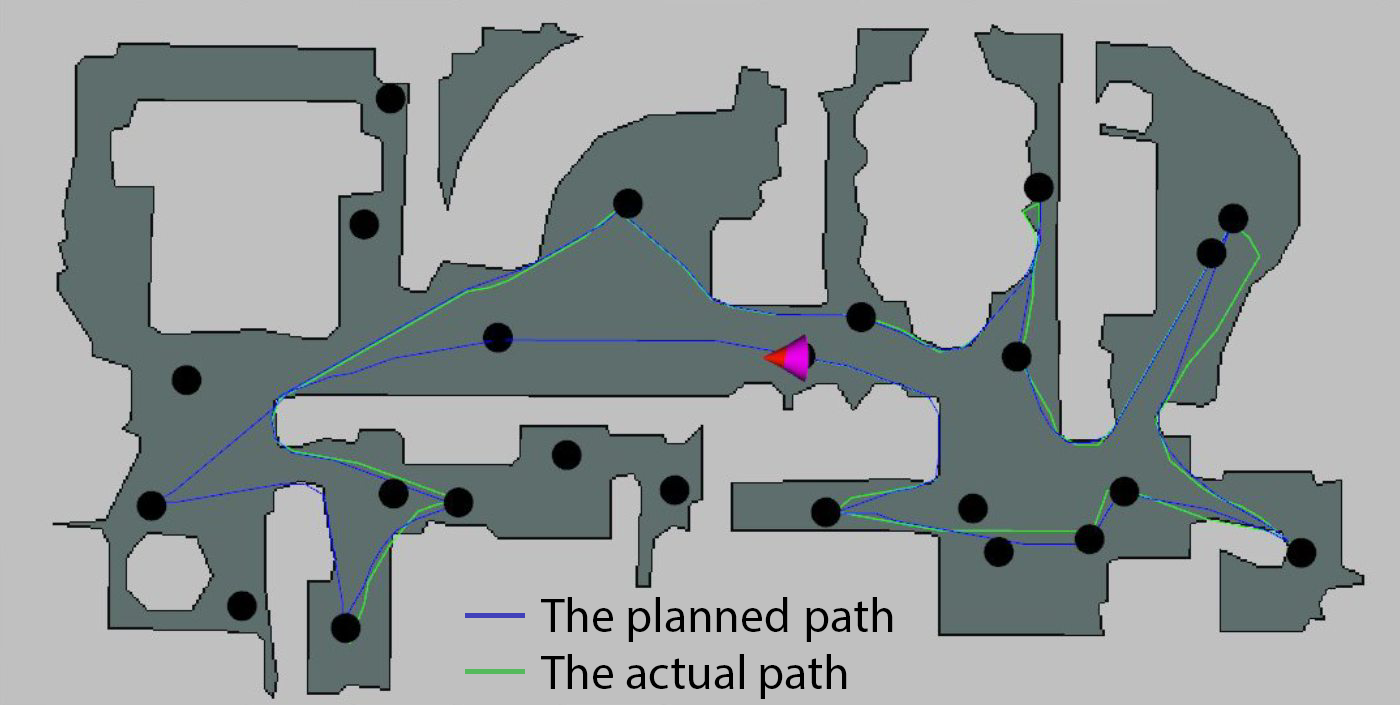}
    \caption{The robot paths in the presence of action uncertainty (10\% random actions, 90\% correct actions).}
    \label{fig:action}
\end{figure}

\begin{figure}[h]
    \centering
    \includegraphics[width=0.9\linewidth]{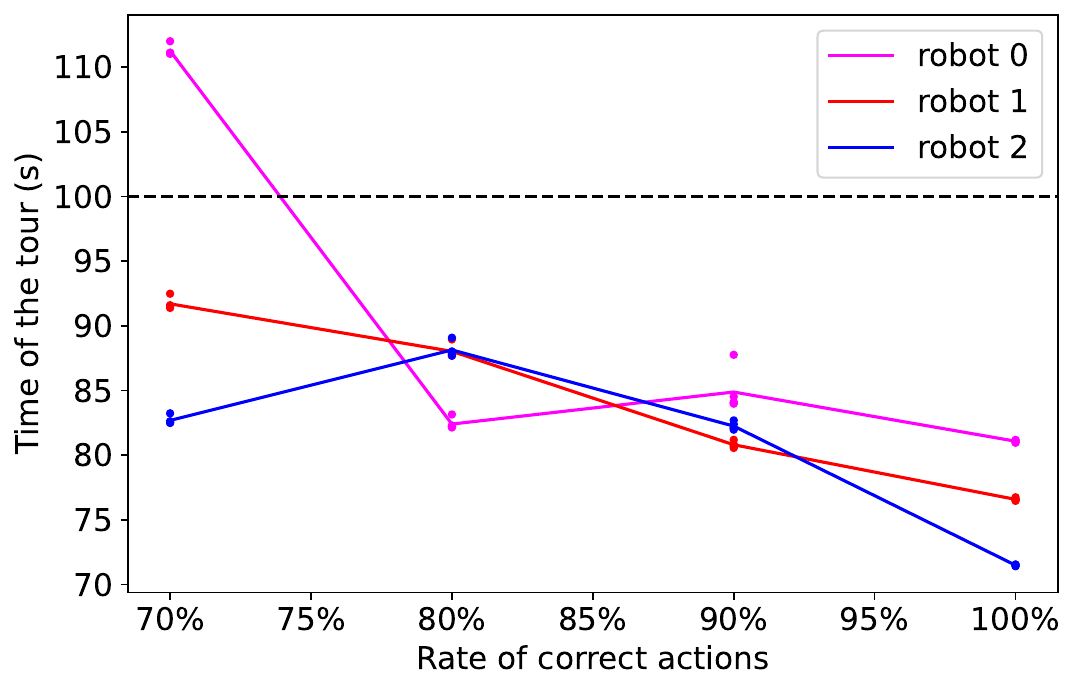}
    \caption{The actual tour time of the three robots when the environmental uncertainty changes.
    The dots show the results of multiple random trials, while the line shows the mean.
    }
    \label{fig:sim_rate}
\end{figure}

\begin{figure}[h]
    \centering
    \includegraphics[width=0.8\linewidth]{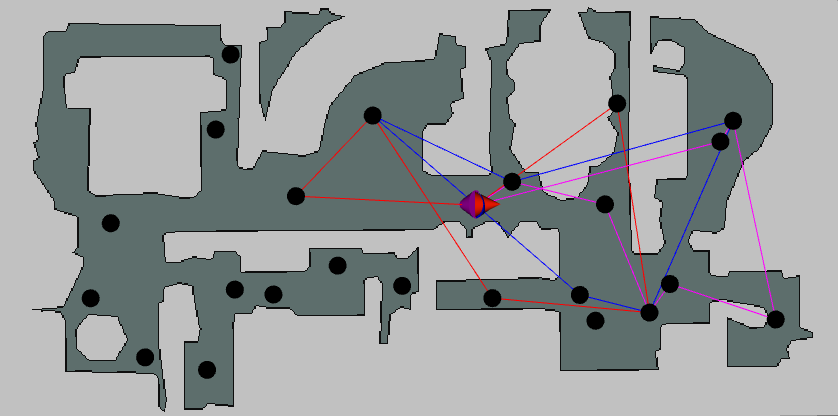}
    \caption{The planned tour (shown as straight lines) of the three robots assuming the standard deviation is 40\%.
    % Between two POIs, straight lines instead of the actual trajectories are shown for clarity.
    }
    \label{fig:sigma4}
\end{figure}

\subsection{Simulation Evaluation}

To evaluate the robustness of the generated plan under practical uncertainties, we complement the tour planner with local navigation planners for each robot and develop a simulation program (Fig. \ref{fig:camera_view}) based on Habitat-AI \cite{savva2019habitat, chang2017matterport3d, kadian2020sim2real}.

\textbf{Uncertainty in the traveling time} \(T_{k i j}\): we manually add \textbf{random actions} to the robot navigator to simulate the situation where robots are disturbed by the environment or when they wait for a human. The effect of 10\% random actions is shown in Fig. \ref{fig:action}. The \textbf{robot dynamics} also add uncertainty to the traveling time.

\textbf{Uncertainty in the visiting time} at each POI \(T_{i j}\): in the simulator, the time is sampled from a Gaussian distribution centered at a manually defined nominal time.

The SMRP planner assumes estimated distributions of time costs are available.
In reality, we can gather samples to estimate the time uncertainty, but the estimation can be inaccurate.
We did a parametric study on the level of environmental and estimated uncertainties to see how planned tours and actual tour times change accordingly.

In Fig. \ref{fig:sim_rate}, the planner assumes that the time uncertainties are Gaussian and the standard deviations are 40\% of the nominal expected values (an inaccurate estimation) and generates a plan (Fig. \ref{fig:sigma4}). These results indicate that the plans can ensure the time constraints even under action uncertainties with a relatively wide margin (when correct actions \(\geq\) 80\%).

\section{Conclusions}\label{sec:conclusion}

This paper presents a multi-robot framework that deals with robotic tour guiding in an environment with uncertain traveling and visiting times.
Computation evaluations are conducted to demonstrate the scalability and optimality of the proposed framework and algorithms.
Parametric studies and simulation evaluations demonstrate that the planner can generate robust tours that keep the time constraints in practice with crude uncertainty estimations.

The SMRP described in Sec. \ref{sec:global_planner} only models the core aspects of a practical tour guide problem.
Additional user-defined practical constraints can be added. Examples include: time window constraints (certain POIs are available only during fixed time windows), sequence dependencies (the visit of certain POIs has prerequisites that other POIs be visited first), and human dependencies (some people may prefer to be assigned to the same robots, such as families and friends).

{\small
\bibliographystyle{IEEEtran}
\bibliography{bib/ieee_full, bib/strings_full, ms}

\begin{thebibliography}{10}
\providecommand{\url}[1]{#1}
\csname url@samestyle\endcsname
\providecommand{\newblock}{\relax}
\providecommand{\bibinfo}[2]{#2}
\providecommand{\BIBentrySTDinterwordspacing}{\spaceskip=0pt\relax}
\providecommand{\BIBentryALTinterwordstretchfactor}{4}
\providecommand{\BIBentryALTinterwordspacing}{\spaceskip=\fontdimen2\font plus
\BIBentryALTinterwordstretchfactor\fontdimen3\font minus
  \fontdimen4\font\relax}
\providecommand{\BIBforeignlanguage}[2]{{%
\expandafter\ifx\csname l@#1\endcsname\relax
\typeout{** WARNING: IEEEtran.bst: No hyphenation pattern has been}%
\typeout{** loaded for the language `#1'. Using the pattern for}%
\typeout{** the default language instead.}%
\else
\language=\csname l@#1\endcsname
\fi
#2}}
\providecommand{\BIBdecl}{\relax}
\BIBdecl

\bibitem{satake2009approach}
S.~Satake, T.~Kanda, D.~F. Glas, M.~Imai, H.~Ishiguro, and N.~Hagita, ``How to
  approach humans? strategies for social robots to initiate interaction,'' in
  \emph{Proceedings of the {ACM}/{IEEE} International Conference on Human-Robot
  Interaction}.\hskip 1em plus 0.5em minus 0.4em\relax San Diego, CA: IEEE,
  2009, pp. 109--116.

\bibitem{shiomi2009field}
M.~Shiomi, T.~Kanda, D.~F. Glas, S.~Satake, H.~Ishiguro, and N.~Hagita, ``Field
  trial of networked social robots in a shopping mall,'' in \emph{Proceedings
  of the {IEEE}/{RSJ} International Conference on Intelligent Robots and
  Systems}.\hskip 1em plus 0.5em minus 0.4em\relax IEEE, 2009, pp. 2846--2853.

\bibitem{doering2019neural}
M.~Doering, T.~Kanda, and H.~Ishiguro, ``Neural-network-based memory for a
  social robot: Learning a memory model of human behavior from data,''
  \emph{ACM Transactions on Human-Robot Interaction (THRI)}, vol.~8, no.~4, pp.
  1--27, 2019.

\bibitem{kirby2010affective}
R.~Kirby, J.~Forlizzi, and R.~Simmons, ``Affective social robots,''
  \emph{Robotics and Autonomous Systems}, vol.~58, no.~3, pp. 322--332, 2010.

\bibitem{burgard1998interactive}
W.~Burgard, A.~B. Cremers, D.~Fox, D.~H\"{a}hnel, G.~Lakemeyer, D.~Schulz,
  W.~Steiner, and S.~Thrun, ``The interactive museum tour-guide robot.''\hskip
  1em plus 0.5em minus 0.4em\relax USA: AAAI, 1998, p. 11–18.

\bibitem{burgard1999museum}
W.~Burgard, A.~B. Cremers, D.~Fox, D.~H{\"a}hnel, G.~Lakemeyer, D.~Schulz,
  W.~Steiner, and S.~Thrun, ``The museum tour-guide robot rhino,'' in
  \emph{Autonome Mobile Systeme 1998}.\hskip 1em plus 0.5em minus 0.4em\relax
  Springer, 1999, pp. 245--254.

\bibitem{de2019social}
B.~De~Carolis, G.~Palestra, C.~Della~Penna, M.~Cianciotta, and A.~Cervelione,
  ``Social robots supporting the inclusion of unaccompanied migrant children:
  Teaching the meaning of culture-related gestures,'' \emph{Journal of
  e-Learning and Knowledge Society}, vol.~15, no.~2, pp. 43--57, 2019.

\bibitem{preferences}
Y.~Kurata, ``Interactive assistance for tour planning,'' in \emph{Spatial
  Cognition VII}, C.~H{\"o}lscher, T.~F. Shipley, M.~Olivetti~Belardinelli,
  J.~A. Bateman, and N.~S. Newcombe, Eds.\hskip 1em plus 0.5em minus
  0.4em\relax Berlin, Heidelberg: Springer Berlin Heidelberg, 2010, pp.
  289--302.

\bibitem{fu2022learning}
B.~Fu, W.~Smith, D.~Rizzo, M.~Castanier, M.~Ghaffari, and K.~Barton, ``Learning
  task requirements and agent capabilities for multi-agent task allocation,''
  \emph{arXiv preprint arXiv:2211.03286}, 2022.

\bibitem{bennewitz2005towards}
M.~Bennewitz, F.~Faber, D.~Joho, M.~Schreiber, and S.~Behnke, ``Towards a
  humanoid museum guide robot that interacts with multiple persons,'' in
  \emph{Proceedings of the {IEEE}-{RAS} International Conference on Humanoid
  Robots}.\hskip 1em plus 0.5em minus 0.4em\relax IEEE, 2005, pp. 418--423.

\bibitem{ford2015flows}
L.~R. Ford and D.~R. Fulkerson, \emph{Flows in networks}.\hskip 1em plus 0.5em
  minus 0.4em\relax Princeton university press, 2015.

\bibitem{cheng1996maximum}
Y.-Q. Cheng, V.~Wu, R.~Collins, A.~R. Hanson, and E.~M. Riseman,
  ``Maximum-weight bipartite matching technique and its application in image
  feature matching,'' in \emph{Visual Communications and Image Processing},
  vol. 2727.\hskip 1em plus 0.5em minus 0.4em\relax International Society for
  Optics and Photonics, 1996, pp. 453--462.

\bibitem{carion2020end}
N.~Carion, F.~Massa, G.~Synnaeve, N.~Usunier, A.~Kirillov, and S.~Zagoruyko,
  ``End-to-end object detection with transformers,'' in \emph{Proceedings of
  the European Conference on Computer Vision}.\hskip 1em plus 0.5em minus
  0.4em\relax Springer, 2020, pp. 213--229.

\bibitem{aggarwal2011online}
G.~Aggarwal, G.~Goel, C.~Karande, and A.~Mehta, ``Online vertex-weighted
  bipartite matching and single-bid budgeted allocations,'' in
  \emph{Proceedings of the {ACM}-{SIAM} Symposium on Discrete
  Algorithms}.\hskip 1em plus 0.5em minus 0.4em\relax SIAM, 2011, pp.
  1253--1264.

\bibitem{dutta2019one}
A.~Dutta and A.~Asaithambi, ``One-to-many bipartite matching based coalition
  formation for multi-robot task allocation,'' in \emph{Proceedings of the
  {IEEE} International Conference on Robotics and Automation}.\hskip 1em plus
  0.5em minus 0.4em\relax IEEE, 2019, pp. 2181--2187.

\bibitem{mendoza2013multi}
J.~E. Mendoza and J.~G. Villegas, ``A multi-space sampling heuristic for the
  vehicle routing problem with stochastic demands,'' \emph{Optimization
  Letters}, vol.~7, no.~7, pp. 1503--1516, 2013.

\bibitem{secomandi2009reoptimization}
N.~Secomandi and F.~Margot, ``Reoptimization approaches for the vehicle-routing
  problem with stochastic demands,'' \emph{Operations research}, vol.~57,
  no.~1, pp. 214--230, 2009.

\bibitem{marinakis2013particle}
Y.~Marinakis, G.-R. Iordanidou, and M.~Marinaki, ``Particle swarm optimization
  for the vehicle routing problem with stochastic demands,'' \emph{Applied Soft
  Computing}, vol.~13, no.~4, pp. 1693--1704, 2013.

\bibitem{fu2022robust}
B.~Fu, W.~Smith, D.~M. Rizzo, M.~Castanier, M.~Ghaffari, and K.~Barton,
  ``Robust task scheduling for heterogeneous robot teams under capability
  uncertainty,'' \emph{IEEE Transactions on Robotics}, vol.~39, no.~2, pp.
  1087--1105, 2022.

\bibitem{sundar2017path}
K.~Sundar, S.~Venkatachalam, and S.~G. Manyam, ``Path planning for multiple
  heterogeneous unmanned vehicles with uncertain service times,'' in
  \emph{Proceedings of the International Conference on Unmanned Aircraft
  Systems}, 2017, pp. 480--487.

\bibitem{chen2014optimizing}
L.~Chen, M.~H. H{\`a}, A.~Langevin, and M.~Gendreau, ``Optimizing road network
  daily maintenance operations with stochastic service and travel times,''
  \emph{Transportation Research Part E: Logistics and Transportation Review},
  vol.~64, pp. 88--102, 2014.

\bibitem{li2010vehicle}
X.~Li, P.~Tian, and S.~C. Leung, ``Vehicle routing problems with time windows
  and stochastic travel and service times: Models and algorithm,''
  \emph{International Journal of Production Economics}, vol. 125, no.~1, pp.
  137--145, 2010.

\bibitem{gomez2016modeling}
A.~G{\'o}mez, R.~Mari{\~n}o, R.~Akhavan-Tabatabaei, A.~L. Medaglia, and J.~E.
  Mendoza, ``On modeling stochastic travel and service times in vehicle
  routing,'' \emph{Transportation Science}, vol.~50, no.~2, pp. 627--641, 2016.

\bibitem{venkatachalam2019two}
S.~Venkatachalam, M.~Bansal, J.~M. Smereka, and J.~Lee, ``Two-stage stochastic
  programming approach for path planning problems under travel time and
  availability uncertainties,'' \emph{arXiv preprint arXiv:1910.04251}, 2019.

\bibitem{venkatachalam2018two}
S.~Venkatachalam, K.~Sundar, and S.~Rathinam, ``A two-stage approach for
  routing multiple unmanned aerial vehicles with stochastic fuel consumption,''
  \emph{Sensors}, vol.~18, no.~11, p. 3756, 2018.

\bibitem{fu2020heterogeneous}
B.~Fu, W.~Smith, D.~Rizzo, M.~Castanier, and K.~Barton, ``Heterogeneous vehicle
  routing and teaming with gaussian distributed energy uncertainty,'' in
  \emph{Proceedings of the {IEEE}/{RSJ} International Conference on Intelligent
  Robots and Systems}.\hskip 1em plus 0.5em minus 0.4em\relax IEEE, 2020, pp.
  4315--4322.

\bibitem{toth2002vehicle}
P.~Toth and D.~Vigo, \emph{The vehicle routing problem}.\hskip 1em plus 0.5em
  minus 0.4em\relax SIAM, 2002.

\bibitem{schreieck2016matching}
M.~Schreieck, H.~Safetli, S.~A. Siddiqui, C.~Pfl{\"u}gler, M.~Wiesche, and
  H.~Krcmar, ``A matching algorithm for dynamic ridesharing,''
  \emph{Transportation Research Procedia}, vol.~19, pp. 272--285, 2016.

\bibitem{aydin2020matching}
O.~F. Aydin, I.~Gokasar, and O.~Kalan, ``Matching algorithm for improving
  ride-sharing by incorporating route splits and social factors,'' \emph{PloS
  one}, vol.~15, no.~3, p. e0229674, 2020.

\bibitem{fu2021simultaneous}
B.~Fu, T.~Kathuria, D.~Rizzo, M.~Castanier, X.~J. Yang, M.~Ghaffari, and
  K.~Barton, ``Simultaneous human-robot matching and routing for multi-robot
  tour guiding under time uncertainty,'' \emph{Journal of Autonomous Vehicles
  and Systems}, vol.~1, no.~4, p. 041005, 2021.

\bibitem{he2018improved}
L.~He, X.~Liu, G.~Laporte, Y.~Chen, and Y.~Chen, ``An improved adaptive large
  neighborhood search algorithm for multiple agile satellites scheduling,''
  \emph{Computers \& Operations Research}, vol. 100, pp. 12--25, 2018.

\bibitem{alinaghian2018multi}
M.~Alinaghian and N.~Shokouhi, ``Multi-depot multi-compartment vehicle routing
  problem, solved by a hybrid adaptive large neighborhood search,''
  \emph{Omega}, vol.~76, pp. 85--99, 2018.

\bibitem{deng2021room}
M.~Deng, B.~Fu, and C.~Menassa, ``Room match: Achieving thermal comfort through
  smart space allocation and environmental control in buildings,'' in
  \emph{Proceedings of the 2021 Winter Simulation Conference. Phoenix, AZ.},
  2021.

\bibitem{deng2023learning}
M.~Deng, B.~Fu, C.~C. Menassa, and V.~R. Kamat, ``Learning-based personal
  models for joint optimization of thermal comfort and energy consumption in
  flexible workplaces,'' \emph{Energy and Buildings}, vol. 298, p. 113438,
  2023.

\bibitem{heller1956extension}
I.~Heller and C.~B. Tompkins, ``An extension of a theorem of dantzig’s,''
  \emph{Linear inequalities and related systems}, vol.~38, pp. 247--254, 1956.

\bibitem{savva2019habitat}
M.~Savva, A.~Kadian, O.~Maksymets, Y.~Zhao, E.~Wijmans, B.~Jain, J.~Straub,
  J.~Liu, V.~Koltun, J.~Malik, D.~Parikh, and D.~Batra, ``Habitat: A platform
  for embodied ai research,'' in \emph{Proceedings of the {IEEE} International
  Conference on Computer Vision}, 2019, pp. 9339--9347.

\bibitem{chang2017matterport3d}
A.~Chang, A.~Dai, T.~Funkhouser, M.~Halber, M.~Niessner, M.~Savva, S.~Song,
  A.~Zeng, and Y.~Zhang, ``Matterport3d: Learning from rgb-d data in indoor
  environments,'' \emph{arXiv preprint arXiv:1709.06158}, 2017.

\bibitem{kadian2020sim2real}
A.~Kadian, J.~Truong, A.~Gokaslan, A.~Clegg, E.~Wijmans, S.~Lee, M.~Savva,
  S.~Chernova, and D.~Batra, ``Sim2real predictivity: Does evaluation in
  simulation predict real-world performance?'' \emph{IEEE Robotics and
  Automation Letters}, vol.~5, no.~4, pp. 6670--6677, 2020.

\end{thebibliography}
}

\end{document}